# OBJECT SORTING USING FASTER R-CNN

Pengchang Chen and Vinayak Elangovan

Division of Science and Engineering, Penn State Abington, PA, USA
pkc5190@psu.edu
Division of Science and Engineering, Penn State Abington, PA, USA
vue9@psu.edu

## ABSTRACT

*In a factory production line, different industry parts need to be quickly differentiated and sorted for further process. Parts can be of different colors and shapes. It is tedious for humans to differentiate and sort these objects in appropriate categories. Automating this process would save more time and cost. In the automation process, choosing an appropriate model to detect and classify different objects based on specific features is more challenging. In this paper, three different neural network models are compared to the object sorting system. They are namely CNN, Fast R-CNN, and Faster R-CNN. These models are tested, and their performance is analyzed. Moreover, for the object sorting system, an Arduino-controlled 5 DoF (degree of freedom) robot arm is programmed to grab and drop symmetrical objects to the targeted zone. Objects are categorized into classes based on color, defective and non-defective objects.*

## KEYWORDS

*Quality inspection, Defects detection, Fast R-CNN, Faster R-CNN,*

## 1. INTRODUCTION

In manufacturing industries, product inspection is an important step in the production process. The quality of products is one of the most important factors of production. Quality Inspection (QI) helps the manufacturing industry determine whether a product is qualified or not. Visual inspection is the key technology of product quality. In the production process, the main purpose of this inspection is to eliminate products that do not meet requirements. The massive repetition nature of QI arises the demand for RPA (Robotic process automation). While the QI of most manufacturing products has a clear standard, it is possible that we can train the robots to classified the qualified and unqualified products, improve the product qualification ratio, and increase productivity.

In the modern production process, a fast, accurate, and precise QI is needed to reduce the inspection cost and also to improve the process efficiency. Manufactured parts gathered from a conveyor are typically auto analyzed to differentiate from defective and non-defective products. These parts are inspected based on features like shape, color, size, texture, etc. Automating the Object sorting process is massively utilized among industries. The traditional object detection process involves three main stages: Informative Region Selection, Feature Extraction, and Classification [1]. At each stage, more image processing is performed to extract information from the images. In the first stage, multiple windows are created to show objects that will possibly appear in the image, which would result in creating needless windows. In the feature extraction stage, as the name implied, features like shape, texture, and size will be extracted from the object in the image. Lastly, to determine which category does the object belongs to, machine learning supervised algorithms, can be used to accomplish the goal.

In generic object detection, the convolutional neural network (CNN) model is the most representative model of deep learning, and several other models are derived from CNN with performance enhanced. For example, region-based full CNN (R-FCN), fast region-based CNN (FRCN), faster region-based CNN (Faster R-CNN) e.tc. Among various industries, with the implementations of the CNN network, the factory can save up to 67% of staff costs and 91.6% of accuracy in detecting target of interests among different objects [2]. Faster R-CNN model is trained with mango and pitaya images and utilized to detect and sort multi-fruits, and as a result, the model achieved 99% of precision in sorting multi-fruits while supporting the quality of fruits [3]. Moreover, the R-FCN model trained with sufficient labeled data has been proven to be efficient in locating and recognizing various objects [4].

Object detection techniques have been widely utilized in different areas. An evaluation and analysis of three deep convolutional neural network architectures to computer-aided detection problems, dataset features (including choosing either large scale of dataset or finer object detection models), and CNN transfer learning from non-medical to medical image field [5]. The three main CNN architectures exploited in [5] are CifarNet, AlexNet, GooLeNet. Another research also conducted a deep neural network to perform detection and classification for multi-fruits in agricultural industries [3]. And the model implemented is a faster region-based convolutional neural network (Faster R-CNN).

Another study applied probabilistic models to detect symmetric objects using images of real-world architectures [e.g., façade, windows, door, shop, balcony e.tc], and the models that research conducted are Weak Structure Model, Spatial Pattern Templates, and Bayesian inference [6]. Moreover, Hsieh and Chen [7] proposed a novel approach that is using symmetrical SURF (speeded-up robust features detector) to search for regions with high vertical symmetry to detect vehicles. Simultaneously, grid division and support vector machine (SVM) learning algorithms are carried out in the research. Another study for detecting and recognizing fasteners' condition (intact, partially worn, and completely missing) on the railway was accomplished by Liu e.tc [8]. The research exploits an algorithm that generates symmetrical samples of fastener images dataset and combines images for enhanced fasteners recognition using an advanced sparse representation algorithm for symmetrical image recognition. Garad [9] researched object sorting based on color, size, and shape identification using MATLAB to find variations in the specific geometric measurements. However, any defects detection on the surface is not addressed.

Motivated by the research done by others, this paper aims to testify the practicality and feasibility of a faster R-CNN model using a dataset containing images of symmetric objects. An auto-sorting system is a setup that integrates output from the neural network for a robotic arm to pick and drop defective objects from a conveyor belt. In this paper, an auto-sorting system is proposed using 5 DoF robot arms with Arduino Mega 2560 Rev3 and two 8-channel DC 5v relay module connected, and a webcam is fixed at a position to take images of objects. Colour detection is implemented using HSV background subtraction and the images are fed to a neural network to perform classification. A faster RCNN model is applied and trained to differentiate defective objects among intact objects based on shape features. Lastly, if a defective cube is detected based on the prediction result, the robot arm is auto programmed to pick and drop the cubes to a designated location. This paper is organized as follows: 2. Neural Network Algorithms, 3. Colour Detection, 4. Hardware of Robotic Arm, 5. Implementation of Faster R-CNN, 6. Results Analysis, 7. Conclusions and References.

## 2. NEURAL NETWORK ALGORITHMS

### 2.1 Architecture of CNN

CNN is a deep learning algorithm, a representative image classification method, that takes images as input, processes them with multiple layers (filters), and classifies objects in images with specific categories as output [e.g., people, cars, animals, materials e.tc.]. The general processing outline is shown in figure 1:

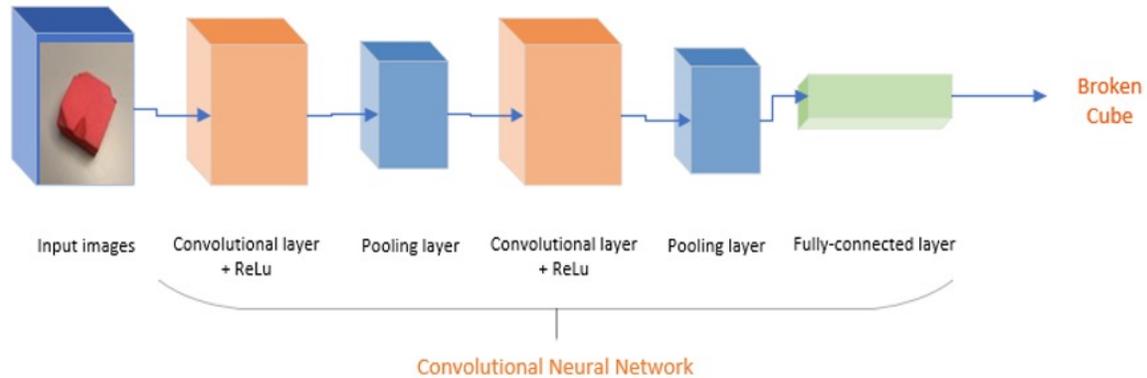

Figure 1. CNN flowchart

The computer sees images as an array of pixels written in a matrix, and there is also pixel matrix representing the features. The common CNN architecture contains three main layers: convolutional layer, pooling layer, and fully-connected layer. Each layer carries out different tasks. The first layer of CNN is the convolutional layer which reduces the image into a form without features loss. During convolution operation, the extracted feature matrix performs dot products with a filter matrix to get a new matrix called a convolved feature. The ReLU layer stands for rectified linear unit, which increases the non-linearity in images by thresholding images at zero. The pooling layer, such as max pooling, average pooling, and sum pooling, are carried out to reduce redundant parameters (spatial size) in a convolved feature that would slow down computation efficiency in the later process. For different types of pooling, it is responsible for different tasks. Max pooling returns a matrix that is made up of the maximum value from the portion of the convolved feature. The logic also applies to the remaining types. Lastly, the fully-connected layer, which is a stage of classification. It summarizes performance from previous layers and returns a result of scores corresponding to object categories.

### 2.2 The architecture of R-CNN and Fast R-CNN

Prior to fast R-CNN, RCNN (region-based convolutional neural network), a successor of the CNN model, partially contributes to the completion of a fast R-CNN model. The general process of R-CNN is shown in figure 2. Different from CNN, RCNN selectively searches to generate proposal images from the input image with arbitrary dimensions at the first stage. Since deep convolutional network requires input data with uniform size, generated region proposals are further warped or cropped into a fixed resolution. After feature extraction, each warped region proposals are fed into a convolutional network and scored based on positive regions and negative regions. The scored regions fit into bounding-box (BBs) regression for final object localization.

Despite outstanding object detection accuracy and classification performed by R-CNN, there are still notable drawbacks: space-consuming and time-consuming in the training process, slow in object detection, and training procedure is a multi-step pipeline [11].

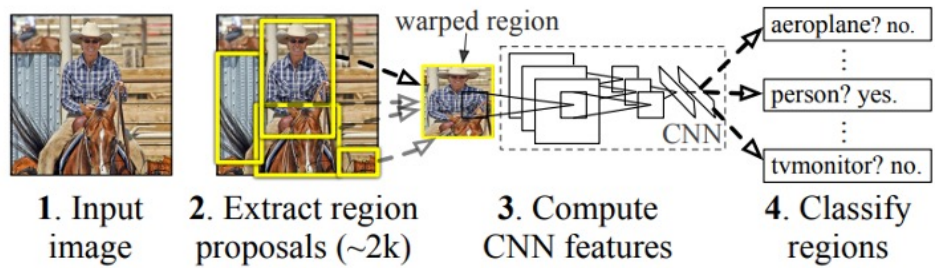

Figure 2. R-CNN flowchart [10]

In fast R-CNN, the novel algorithm presents an update to RCNN and SPPnet, and it achieves 9 times faster on training VGG16 network (a CNN architecture) than R-CNN and 213 times faster on testing. The flow of fast R-CNN is shown in figure 3. In the procedure of fast R-CNN, it receives an entire image as input and a group of region proposals. Collected images are passing through a multilayer Conv net and pooling layer to form a feature map. And for each region proposals, a fixed-length feature vector is extracted by the region of interests (RoI) pooling layer. A series of fully-connected layers receive feature vectors to produce two outputs: one with softmax probability and the other yields 4 numbers that represent the location of the bounding box for each detected object.

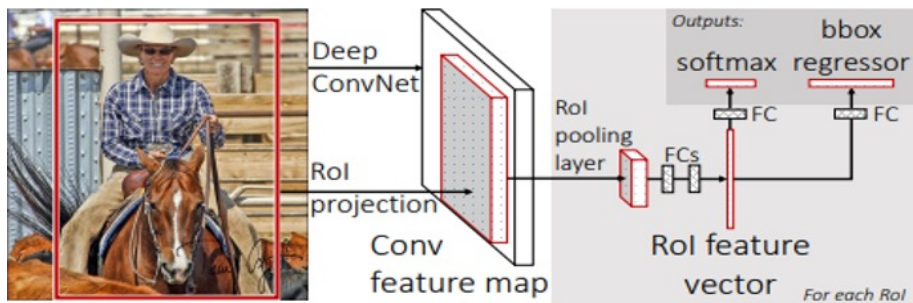

Figure 3. Fast R-CNN flowchart [11]

## 2.3 The architecture of Faster R-CNN

State-of-art object detection networks mainly use region proposal algorithm to assume object position. R-CNN and fast R-CNN has accomplished to reduce the training and detecting time considerably by applying region proposal algorithm, there is still a bottleneck in improving region proposal computation efficiency. To address this problem, Ren et al. [12] proposed region proposal network (RPN) that allows nearly cost-free computation in generating region proposals.

In the architecture of faster R-CNN, there are two main modules covered, RPN and Fast R-CNN object detector. The flowchart of faster R-CNN is shown in figure 4. Images are passed through convolutional layers to generate a feature map, where RPN collects processed images with random size (avoid object shape deformed due to rescaling) to predict a group of objects with objectness scores (determine if it's an object or not). Fixed-length feature vectors are extracted from each proposal using RoI pooling. Feature vectors are then inserted into a series of fully-connected layers with two sibling outputs. The system returns a detection of the target object and the other output returns 4 real numbers that represent the location of the bounding box around the detected object.

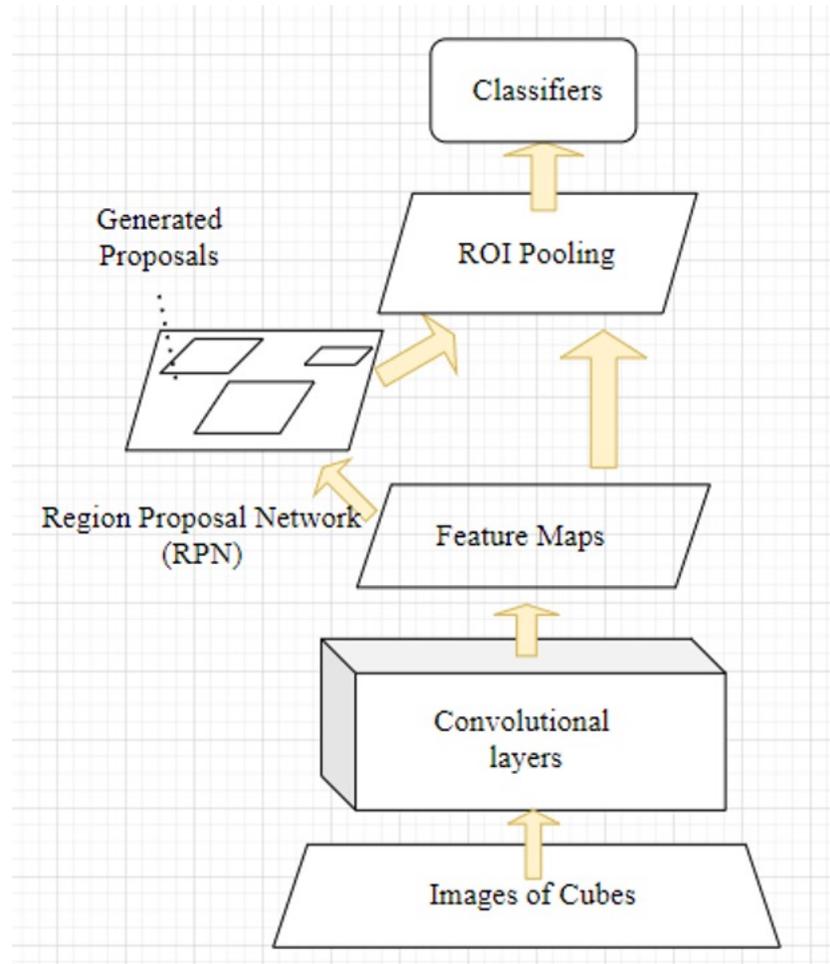

Figure 4. Faster R-CNN flowchart

## 2.4 Selection of Algorithms

The applications of CNN to many regions based neural networks have achieved significant results as well as notable drawbacks. Both R-CNN and Fast R-CNN implement a selective search algorithm for generating region proposals. Nevertheless, a high computation process results in space-consuming and time-consuming problems and slow in detection time. Faster R-CNN is the aggregation of R-CNN and fast R-CNN algorithms, which is used to resolve image classification problems. Simultaneously, performance carried out by faster R-CNN indicates a prominent improvement in PASCAL VOC 2007 and 2012 test dataset. The detection time is 250 times faster than R-CNN and 10 times faster than Fast R-CNN as shown in table 1. The references for the mean average precision rate are also shown in square brackets in table 1.

As the promising result achieved by faster R-CNN, it motivated us to explore an implementation of faster R-CNN to an auto sorting system. And it is expected that the computation process will be faster and better accuracy.

Table 1. Comparison Between Different R-CNN

|  | Mean Average Precision | | |
| --- | --- | --- | --- |
|  | R-CNN | Fast R-CNN | Faster R-CNN |
| PASCAL VOC 2007 Test Dataset (%) | 58.5[10] | 66.9 [11] | 69.9 [12] |
| PASCAL VOC 2012 Test Dataset (%) | 53.3[10] | 65.7 [11] | 67.0 [12] |
| PASCAL VOC 2007 + PASCAL VOC 2012 Test Dataset (%) | - | 70.0 [11] | 73.2 [12] |
| PASCAL VOC 2007 + PASCAL VOC 2012 + COCO Test Dataset (%) | - | - | 78.8 [12] |
| Prediction Time / image (seconds) | ~50 sec | ~2 sec [11] | ~0.2 sec [12] |

## 3. COLOUR DETECTION

To detect colored intact cubes and defect cubes, we accomplished the goal by using OpenCV (open-source computer vision library) and python 3.7. The color detection is achieved by capturing images of cubes and classified by its color type BGR. The idea of the algorithm is to change the color space of each frame to HSV (Hue, Saturation, Value) color space while the fixed camera is capturing the images. Then we define a range of values that specify light and dark color in the HSV scale. In HSV colour space in OpenCV, the value range for three matrixes are 0-179, 0-255, 0-255 respectively. Then we threshold the HSV frames to detect colored objects that have HSV value defined in the range. Finally, a green rectangle box will be generated and drawn around the detected colored cubes for testing purposes as shown in figure 5. After each colored cube is detected, a picture (without rectangle box) will be stored in a folder for neural network prediction usage.

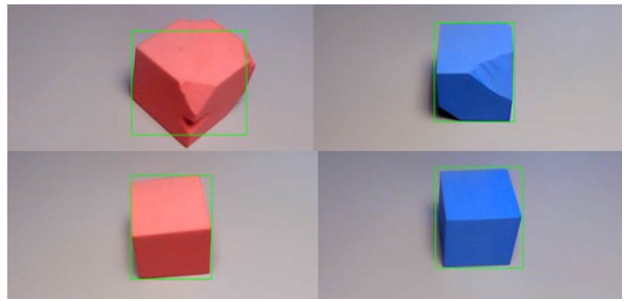

Figure 5. Result of color detection

## 4. HARDWARE OF ROBOT ARM

The auto sorting system contains a 5 DoF robot arm with DC motors, battery, Arduino Mega 2560 Rev3 and two 8-channel DC 5v relay module and a dozen of female-to-male jumper wires. The robotic arm shown in Figure 6 is OWI Robotic Arm Edge. It has 5 degrees of freedom: a gripper that can be opened or closed, a wrist motion of 120 degrees, an extensive elbow range of 300 degrees, base rotation of 270 degrees, and base motion of 180 degrees. It can lift objects up to 100g, reach 15 inches vertically, and 12.6 inches horizontally. It has a no-load current of 255mA. Four type D batteries are required to provide the current and a voltage of 3V. The robotic arm is controlled using an Arduino Board, which is a low-cost, open-source microcontroller. While many microcontrollers are available in the market, such as ROS, LinuxCNC, and raspberry, Arduino is a more cost-effective choice. The Arduino codes developed in Arduino IDE are uploaded to the Arduino microcontroller to control the current flow out of the pins or read the current flow of the pins.

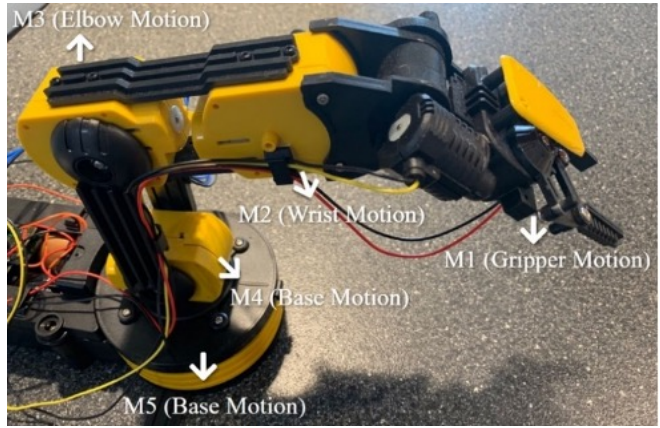

Figure 6. Robot Arm

## 5. IMPLEMENTATION OF FASTER R-CNN

To speed up the training computation process, image rescaling is required. Each image is rescaled to 50% of its original size. Moreover, labeling for input images is done before feeding data into faster R-CNN. We use an application called *labelimg* to generate RoI and select a region where object features are covered and we further label regions with proper category, respectively. The labeling process is shown in figure 7.

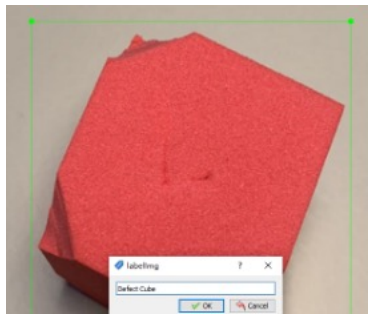

Figure 7. Defected Object - Image labeling

After each input images are labeled and stored as xml files, we will get hyperparameters H x W ( H = height, W = width) and four number representing selected region coordinates. To facilitate the image process, we convert our xml files to csv files, figure 8. The collected images of defect and intact cubes and labels are then fed into a faster R-CNN model using Keras open-source neural network library and Tensorflow open source library.

|   | filename | width | height | class | xmin | ymin | xmax | ymax |
|---|----------|-------|--------|-------|------|------|------|------|
| 0 | defect0.jpg | 540 | 610 | defect | 64 | 92 | 352 | 322 |
| 1 | defect1.jpg | 540 | 610 | defect | 66 | 61 | 352 | 340 |
| 2 | defect10.jpg | 540 | 610 | defect | 69 | 63 | 323 | 323 |
| 3 | defect12.jpg | 540 | 610 | defect | 53 | 69 | 351 | 336 |
| 4 | defect13.jpg | 540 | 610 | defect | 67 | 60 | 328 | 326 |

Figure 8. xml to csv

Each image is passed through hidden layers in the model during the computation process. As a result, a faster R-CNN model achieved 99% accuracy in detecting defect cube and 97% in detecting intact cube, as result shown in figure 9. However, the accuracy is fluctuating for some cases. To maintain the higher accuracy for other cases, a large scale of dataset with correct labeling would contribute to a steady and promising result

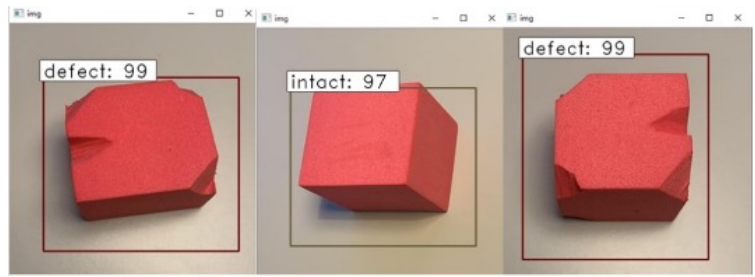

Figure 9. Result of Prediction

## 6. RESULT ANALYSIS

In this section, we analyze the performance of a faster R-CNN algorithm in the auto sorting system using confusion matrix. Each formation of confusion matrix is based on the size of the training dataset and testing dataset. We exploit 400 images of intact cubes and defective cubes as dataset, and we pass different percentages of the dataset into the neural network model and record the prediction results shown in the tables below.

Table 2. 70% Training Dataset and 30% Testing Dataset

| N = 120 (35 intact, 85 defect) | Intact Predicted (negative) | Defects Predicted (positive) |   |
|---|---|---|---|
| Actual Intact (False) | 24 (TN) | 17(FP) | 41 |
| Actual Defects (True) | 9 (FN) | 65 (TP) | 79 |
|   | 33 | 87 | N = 120 |

During the training process, we implemented 5 epochs with epoch length equivalent to 500, the estimated training time for each epoch is 1680 sec. After the training process is done, we employ different sizes of test images based on the percentage of the training dataset. The accuracy and precision percentages are calculated based on the result from prediction tables including true-positive value, true-negative value, false-positive value, and false-negative value shown in table 5.

Table 3. 80% Training Dataset and 20% Testing Dataset

| N = 80 (28 intact, 52 defect) | Intact Predicted (Negative) | Defects Predicted (Positive) | |
|---|---|---|---|
| Actual Intact (False) | 21 (TN) | 10 (FP) | 33 |
| Actual Defects (True) | 6 (FN) | 43 (TP) | 47 |
| | 27 | 53 | N = 80 |

Table 4. 90% Training Dataset and 10% Testing Dataset

| N = 40 (13 intact, 27 defect) | Intact Predicted (Negative) | Defects Predicted (Positive) | |
|---|---|---|---|
| Actual Intact (False) | 9 (TN) | 6 (FP) | 15 |
| Actual Defects (True) | 0 (FN) | 25 (TP) | 25 |
| | 5 | 35 | N = 40 |

Table 5. Accuracy and Precision

| | 70% Training Dataset and 30% Testing Dataset | 80% Training Dataset and 20% Testing Dataset | 90% Training Dataset and 10% Testing Dataset |
|---|---|---|---|
| Accuracy | 78.83% | 80% | 85.5% |
| Precision | 80.45% | 81.13% | 88.42% |

As the result shown in table 5, with the size of the training dataset increases, the accuracy and precision rate are steadily enhancing. If we enrich the quantity of the training dataset, better promising accuracy, and precision results would be expected.

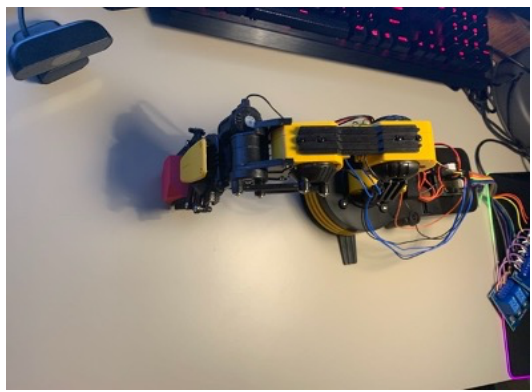

Figure 10. Robotic Arm Picking An Object

After the color detection process and neural network database are accomplished and generated, then the robot arm is initiated to perform pick and drop tasks based on the captured image, whether containing the defective or intact cubes, while matching images to the database for prediction result. Figure 10 illustrates the motion of the robot arm performing pick and drop tasks.

The accuracy rate of detecting defective and non-defective objects is improved successfully when increasing the training dataset. Nevertheless, there are limitations in the training dataset. Image acquisition is an important step for defects detection using image processing algorithms. 400 images of defective and non-defective objects in categorized color are taken at a fixed filming and lighting angle with a plain background to reduce unwanted noises. Any changes in background would result in slower training time and poorer accuracy rate in the detection stage. The difference in choosing the texture of the symmetrical object may also influence the training result of the test result, and the impact is shown on the reflection rate of the images.

## 7. CONCLUSIONS

The research employed a neural network algorithm, faster R-CNN, to build an auto sorting system and compared with R-CNN and fast R-CNN algorithm. Colour detection is done using python 3.7 OpenCV library. The image classification process is accomplished by applying captured images to a pre-trained faster R-CNN model for prediction. As the analysis of how the training dataset impacts the results, we evidenced that if we enrich the dataset of defective and intact cubes, a better prediction result can be expected. The implementation of color detection and pre-trained faster R-CNN model on a 5 DoF robot arm with Arduino Mega and two 8-channel DC 5v relay module-controlled works effectively on executing pick and drop tasks.

In future, this research will be continued with different materials: metallic, plastic and wooden to demonstrated influence of materials in training and detection stages using faster R-CNN as well as YOLOv4 neural network.

## Authors

Dr. Vinayak Elangovan is an Assistant Professor of Computer Science at Penn State Abington. His research interest includes computer vision, machine vision, multi-sensor data fusion and activity sequence analysis with keen interest in software applications development and database management. He has worked on number of funded projects related to Department of Defense and Department of Homeland Security applications.

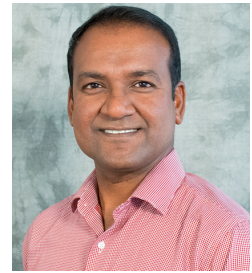

Pengchang Chen is a current junior student at Penn State University and major in computer science and minor in mathematics. His research interests include machine vision, artificial intelligence, robotics. He has participated in multiple projects including encryption and decryption application, augmented reality tool.

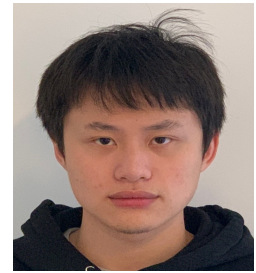